\documentclass[10pt,twocolumn,letterpaper]{article}

\usepackage[pagenumbers]{cvpr} 

\usepackage{graphicx}
\usepackage{amsmath}
\usepackage{amssymb}
\usepackage{booktabs}

\usepackage[pagebackref,breaklinks,colorlinks]{hyperref}

\usepackage[capitalize]{cleveref}
\crefname{section}{Sec.}{Secs.}
\Crefname{section}{Section}{Sections}
\Crefname{table}{Table}{Tables}
\crefname{table}{Tab.}{Tabs.}

\begin{document}

\title{One-shot recognition of any material anywhere using contrastive learning with physics-based rendering}

\author{
Manuel S. Drehwald\textsuperscript{3,*,\#}, Sagi Eppel \textsuperscript{1,2,4,*,\#}, Jolina Li \textsuperscript{2,4}, Han Hao\textsuperscript{2}, Alan Aspuru-Guzik\textsuperscript{1,2,\#}\\
   }

\maketitle

\let\thefootnote\relax\footnotetext{
\textsuperscript{1}Vector institute, \textsuperscript{2}University of Toronto, \\
\textsuperscript{3}Karlsruhe Institute of Technology, \textsuperscript{4}Innoviz \\
* Equal Contributions, \# Corresponding authors \\
alan@aspuru.com, manuel@drehwald.info, sagieppel@gmail.com
\medskip
}

\maketitle

\maketitle

\begin{abstract}
Visual recognition of materials and their states is essential for understanding most aspects of the world, from determining whether food is cooked, metal is rusted, or a chemical reaction has occurred. However, current image recognition methods are limited to specific classes and properties and can't handle the vast number of material states in the world. 
To address this, we present MatSim: the first dataset and benchmark for computer vision-based recognition of similarities and transitions between materials and textures, focusing on identifying any material under any conditions using one or a few examples. The dataset contains synthetic and natural images. The synthetic images were rendered using giant collections of textures, objects, and environments generated by computer graphics artists. We use mixtures and gradual transitions between materials to allow the system to learn cases with smooth transitions between states (like gradually cooked food). We also render images with materials inside transparent containers to support beverage and chemistry lab use cases. 
We use this dataset to train a siamese net that identifies the same material in different objects, mixtures, and environments. The descriptor generated by this net can be used to identify the states of materials and their subclasses using a single image. 
We also present the first few-shot material recognition benchmark with images from a wide range of fields, including the state of foods and drinks, types of grounds, and many other use cases. We show that a net trained on the MatSim synthetic dataset outperforms state-of-the-art models like Clip on the benchmark and also achieves good results on other unsupervised material classification tasks. Dataset and trained models have been made available at these URLS: \href{https://zenodo.org/record/7390166#.Y_6cNIBBxH4}{1} \href{https://e1.pcloud.link/publink/show?code=kZIiSQZCYU5M4HOvnQykql9jxF4h0KiC5MX}{2}
\textbf{}
\end{abstract}
\begin{figure}[t]
  \centering
   \includegraphics[width=0.9\linewidth]{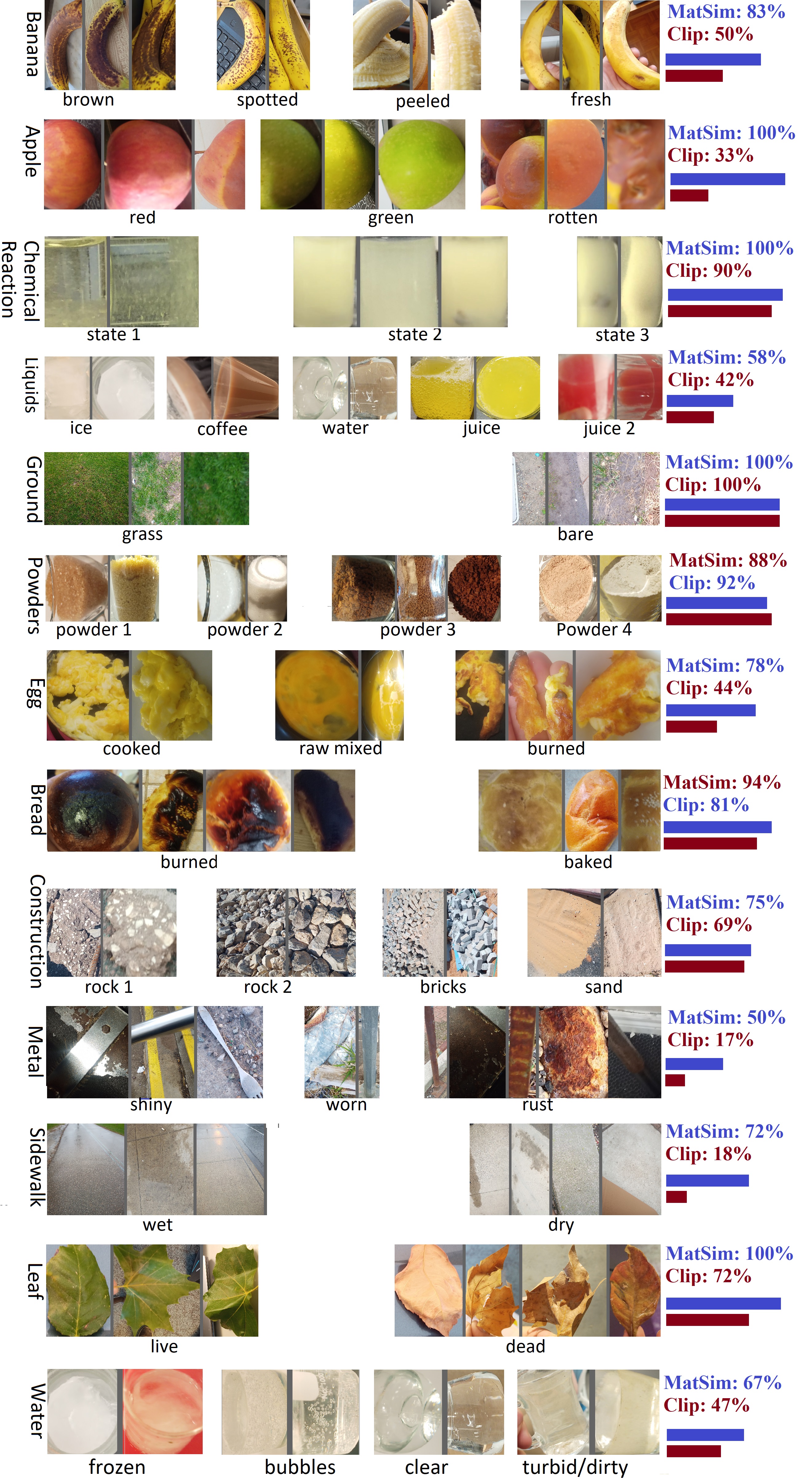}
   \caption{The MatSim benchmark for identifying materials from every aspect of the world using one or a few real images (one shot). Top-1 results of ConvNeXt trained on the synthetic MatSim dataset and Clip H14. Neither of the networks was trained with these classes. Only selected samples are shown.}
    \label{fig:accExamples}
\end{figure}

\begin{figure*}[t]
  \centering
   \includegraphics[width=0.9\linewidth]{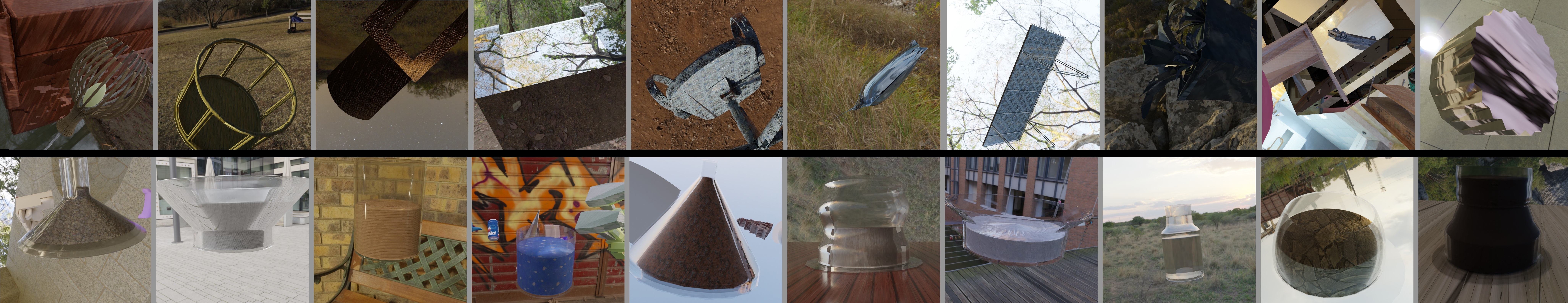}

   \caption{Selected examples from the MatSim synthetic dataset. The top line contains random materials on random objects. The bottom row contains random materials inside transparent vessels.}
   \label{fig:genExamples}
\end{figure*}

\section{Introduction}
\label{sec:intro}

The ability to visually identify materials is critical for a wide range of applications, from material science and chemical research to cooking, construction, and industry (\Cref{fig:accExamples}) \cite{Sun2022using,Zhang2021computer,patricio2018computer,zheng2023materobot,Eppel2020computer,eversberg2021generating}. Being able to recognize materials and their states is essential for handling and inspecting materials in the chemistry lab, evaluating whether the ground is wet, determining whether a fruit is ripe, detecting rust on metal surfaces, and distinguishing between different types of rocks or fabrics. There are several major challenges that make this problem difficult for computer vision methods. First, there is an almost infinite amount of material sates and textures, and each can look very different in different settings. The second challenge is that transitions between material states tend to be gradual and have a continuous intermediate state, which makes it hard to use discrete categories to describe the material (like cooked food). Another challenge is that liquids and materials in environments like laboratories, hospitals, and kitchens are usually handled inside transparent vessels that distort the view of the materials. Previous studies on image recognition for materials have focused on distinguishing between material classes, like metal, plastic, and wood, or determining properties like turbidity or glossiness. The limitation of these approaches is that they can only work the classes and properties they were trained on \cite{dana1999reflectance, bell2015material, hayman2004significance, sharan2009material,  xiao2018unified, Eppel2020computer, bell2013opensurfaces}. To our knowledge,  the problem of identifying any material state in any environment using one example (one-shot learning) has not been addressed by any benchmark or dataset. In this work, we propose the first general dataset and benchmark for one-shot recognition of any material state in any environment using only one or a few examples.

\subsection{The MatSim Dataset}
\label{sec:matsim_int}

The MatSim dataset includes a large-scale synthetic dataset (\Cref{fig:genExamples}) for materials similarity and a diverse real image benchmark for testing the net's ability to identify unfamiliar material states or subclasses using one or a few images (\Cref{fig:accExamples}). The dataset aims to address the general problem of one-shot material retrieval without limitations based on specific material types, environments, or objects. The main focus is on distinguishing between states of materials and identifying fine-grained categories, such as rotten vs ripe or coffee vs cocoa. Additionally, we created a second adversarial benchmark to test the net's ability to recognize materials without association with objects or environments. This benchmark involves covering objects with random materials to create uncorrelated material-object associations (\cref{fig:set2},appendix).

\subsection{Synthetic Dataset and Training}
\label{sec:isynt_int}
Our hypothesis is that training a siamese net to identify the same material texture on different objects and environments will enable the net to recognize material states and subclasses in general. While material states are not always matched to a single texture, we assume that a diverse enough training set will force the net to learn a general representation of materials and their properties. The main advantage of this self-similarity approach is that when applied to synthetic data, it can be used to generate an unlimited amount of data with no human effort. In addition, it can be easily expanded to materials mixtures and gradual transitions. The main challenge is the need for a large and highly diverse dataset to prevent the net from overfitting specific materials or environments.

\subsection{Evaluation and Results}
\label{sec:int_eval}
The net trained on the MatSim dataset achieves good results in recognizing and matching materials of the same states and subclasses, outperforming state-of-the-art nets like CLIP and nets trained on human annotated similarity metrics. We also demonstrate that the net performs well in matching images of the same general class in a standard classification dataset like OpenSurface\cite{bell2013opensurfaces} and DMS\cite{upchurch2022dense}, without using the semantic class labels.
 
\section{Related work}

\subsection{One-Shot and Contrastive Learning}
\label{sec:related_contrastive}
Recent years have seen the emergence of powerful one-shot embedding models like Clip, which rely on contrastive learning to learn image and text similarities\cite{radford2021learning, schuhmann2022laion}.
The main advantage of these approaches is that when trained on enough data, they are not limited to a specific set of cases and can work with new examples and classes \cite{radford2021learning,hendrycks2021many}. These nets work by predicting a descriptor vector for an image and using the distance between different image descriptors as a similarity metric. The limitation of this approach is that it requires an enormous amount of training data, as seen in the Laion Clip H14 model, which was trained on two billion images and their text labels \cite{schuhmann2022laion}. An alternative approach, like SimClr\cite{chen2020simple}, uses self-similarity between images and their augmented versions and does not rely on human-made semantic labels. As a result, this approach has an unlimited amount of training data \cite{chen2020simple}.  This work takes a similar approach and assumes that by learning to identify the same material in different physical settings and mixtures, the net will learn a general descriptor for materials and their states.

\subsection{Image Retrieval and Materials Similarity}
\label{sec:related_retrieval}

The problem of one-shot material recognition and material matching has been mostly ignored from a computer vision perspective. However, it is closely related to the problem of image retrieval and similarity, which involves finding a similar example from a set. Two such approaches were suggested to identify the visual similarity of simulated materials for CGI material replacement. Schwartz and Nishino\cite{schwartz2019recognizing} trained a network to measure the similarity of materials using 9,000 generated images with uniform materials. The visual similarity between the simulated materials was determined by human annotators. Perroni et al.\cite{perroni2022material} used transfer learning to determine the visual similarity by using the inner layer of a network trained on classification tasks with eight classes of materials. While both of these works generate material descriptors that can be used to assess similarity, both have been limited to a specific domain of CGI materials swapping and were not intended for materials recognition. As a result, the human-guided net\cite{schwartz2019recognizing} performed poorly on all real-world material recognition tasks we tested (the others have not been made available\cite{perroni2022material}). 

\subsection{Computer Vision for Materials}
\label{sec:related_vision}

Computer vision for materials has been primarily focused on a few problems: Materials segmentation which involves finding the region of the image belonging to specific materials. Inverse rendering, which involves extracting textures maps from object surfaces (For CGI purposes). Predicting values of specific properties of the material, like glossiness or turbidity\cite{chadwick2015perception, weiss2020correspondence, meka2018lime, vidaurre2019brdf,weiss2020correspondence,lagunas2019similarity, schwartz2019recognizing}. Neither of these areas where intended is used for one-shot material recognition or can be directly applied to this problem.
\subsection{Materials Classification}
Materials classification involves assigning a class for material in the image from a given set of categories in the training sets. Datasets for general material classes (wood, metal, glass)  include CURet\cite{dana1999reflectance}, Flicker-Materials FMD\cite{sharan2009material}, KTH-TIPS \cite{hayman2004significance}, MINC\cite{bell2015material} and large scale, diverse datasets like  OpenSurfaces\cite{bell2015material, bell2013opensurfaces}, and Dense Material Segmentation(DMS)\cite{upchurch2022dense}. Other datasets focus on more specific properties, such as material phases (liquids, solids, powders, foam)\cite{Eppel2020computer}, and more specific topics, such as construction materials \cite{brilakis2006construction}, soil type \cite{srivastava2021comprehensive}, crystallization, turbidity\cite{pizzuto2022solis}, etc\cite{mungofa2018chemical, Sun2022using, Zhang2021computer, patricio2018computer,  eversberg2021generating}. The limitation of Nets trained on these datasets is that they are limited to the specific classes in the dataset and cannot work on new classes without specific training.


\section{The MatSim Synthetic Dataset Generation}

The goal of the MatSim dataset is to train a computer vision system capable of recognizing any visually distinguished material state and texture on any surface in any reasonable environment.  The visual appearance of a material depends on its physical properties but also on the object's shape, environment, and illumination. If during training, any of these properties are restricted in some way (for example, only indoor scenes are used) or correlated with other properties (for example, wood materials only appear on trees), the net is unlikely to achieve true generalization for material recognition \cite{hendrycks2021many, barragan2022adversarially}.

\subsection{Generation Procedure}
\begin{figure}[t]
  \centering
   \includegraphics[width=0.9\linewidth]{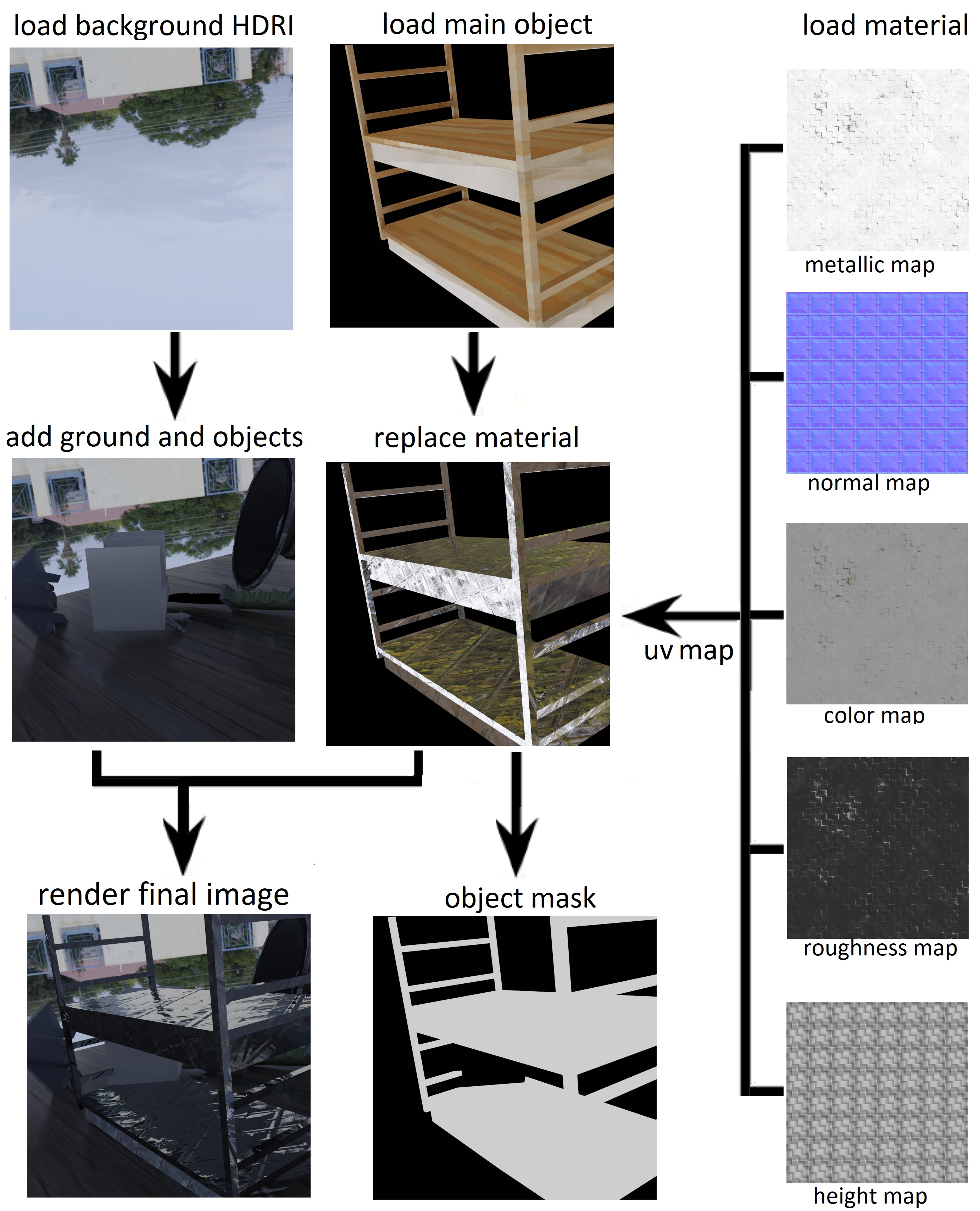}

   \caption{Dataset creation: 1) CGI Materials have been randomly created or downloaded from large-scale artist repositories (AmbientCG\cite{ambientcg}, CGBookCases\cite{cgbookcase}, 3D Textures). 2) The material is UV mapped on the surface of a random object loaded from the ShapeNet dataset\cite{chang2015shapenet}.3) Random background and illumination are loaded from the HDRI Haven repository\cite{polyhaven}. 5) Ground plane and background objects are added. 6) and the scene is rendered.}
   \label{fig:genSteps}
\end{figure}

In order to achieve the above goals, the dataset needs to be extremely diverse in terms of objects, environments, and materials. To achieve this, we utilize large-scale CGI artist repositories\cite{ambientcg, cgbookcase} used for animation and computer games.  We   use thousands of highly diverse physics-based rendering (SVBRDF/PBR) materials repositories to simulate realistic materials\cite{Pharr2016physically}. We overlay the textures materials on 3D objects taken from the ShapeNet dataset with tens of thousands of different objects \cite{chang2015shapenet} and hundred of categories; these objects are then put into scenes with random illumination and backgrounds by utilizing PolyhHaven repository for HDRI images\cite{polyhaven}. The HDRI image is wrapped around the scene and provides a realistic 360-degree background and illumination to the scene (\Cref{fig:genSteps}). 
Combining these repositories allows us to generate a large-scale, highly diverse dataset (\Cref{fig:genSteps}). The large number of materials forces the network to generalize instead of identifying only specific classes. The fact that every material can be used on any object in any environment means that the net has to identify the material everywhere and prevents the net from associating the material with specific objects or environments. Gradual transformations between materials in the dataset allow the net to detect gradual transitions between materials. Additionally, rendering some of the materials inside transparent containers allows the network to learn to recognize materials stored inside glass vessels.
   
\begin{figure}[t]
  \centering
   \includegraphics[width=0.9\linewidth]{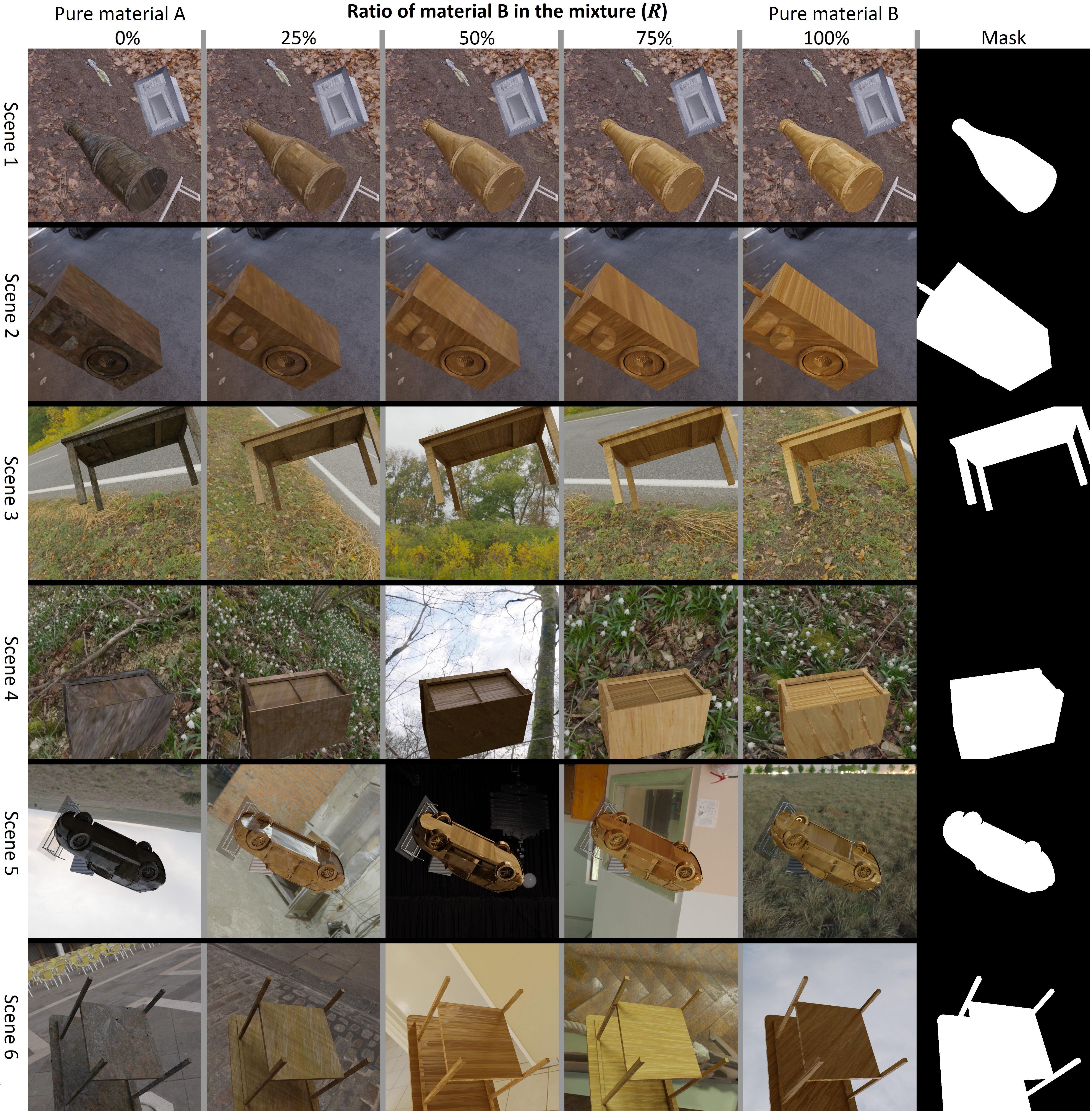}
   \caption{Dataset structure. The dataset is composed of sets. Each set involves two materials and six scenes with a gradual transition between the two materials. Each image column corresponds to a different mixing ratio ($R$) of the two materials. The ratio of the mixture of the two materials is given in the top column. All images in the same column involve the same material mixture on different objects and in different environments. Each of the six scenes involves one main object. The material on this object gradually transitions from one material to another. The mask of the object is given in the right column. For scenes 1–2, the background remains exactly the same for all images in the scene. For scenes 3–4, the background HDRI is randomly rotated between images, leading to small changes in illumination. For scenes 5–6, the background HDRI is completely replaced between images, leading to large changes in illumination.}
   \label{fig:dataSets}
\end{figure}

\subsection{General Dataset Structure} \label{sec:genDataset}
The dataset is divided into sets; each set contains two random materials and six scenes involving a gradual transition between these two materials (\Cref{fig:dataSets}). The objects, backgrounds, and environment are randomly selected for each scene separately (\Cref{fig:genSteps}). Each scene involves one static main object and a static camera, which are both positioned randomly, and both remain unchanged for the scene. The object's material is gradually changed between images in the scene (\Cref{fig:dataSets}). For each scene, we render five images with different mixtures of the two materials. The mixture ratios ($R$) of the two materials are 0\%, 25\%, 50\%, 75\%, and 100\%. With 0\% means that the object is only made of material $A$, while 100\% means the object is made completely of material $B$, and 50\% indicates an equal mixture of the two materials (\Cref{sec:mixtures}).
For scenes 1–2, all aspects of the scene remain the same. Only the object material is changed gradually between images. In addition, the translation and rotation of the texture UV mapping to the object were randomly changed between each render (\cref{sec:apuvmap}). For scenes 3–4, the procedure is the same, except that the HDRI background is randomly rotated between images (\cref{sec:aphdri}), leading to a small variation in illumination. For scenes 5–6, the HDRI background for each image is randomly replaced, leading to a large variation in illumination. This provides almost any possible variation of each material's appearance. Since all the scenes in a set are composed of the same two materials, it is possible to compare the appearance of the materials between two scenes with different objects and backgrounds, and light. The gradual transition allows the network to learn to distinguish between highly similar materials. Some scenes contain several random objects and a ground plane for variability (\cref{sec:apsetobjinscene,sec:apglane}). Since a scene can contain many background objects, for each scene, we provide the mask (region) of the object on which the material is used (\Cref{fig:dataSets}, right). If the material is uniform (BSDF), the values of the material properties (color, transparency, etc.) are given in the dataset. Altogether, 30k sets, with about one million images, were rendered. For more details, see the appendix.
\begin{figure}[t]
  \centering
   \includegraphics[width=0.9\linewidth]{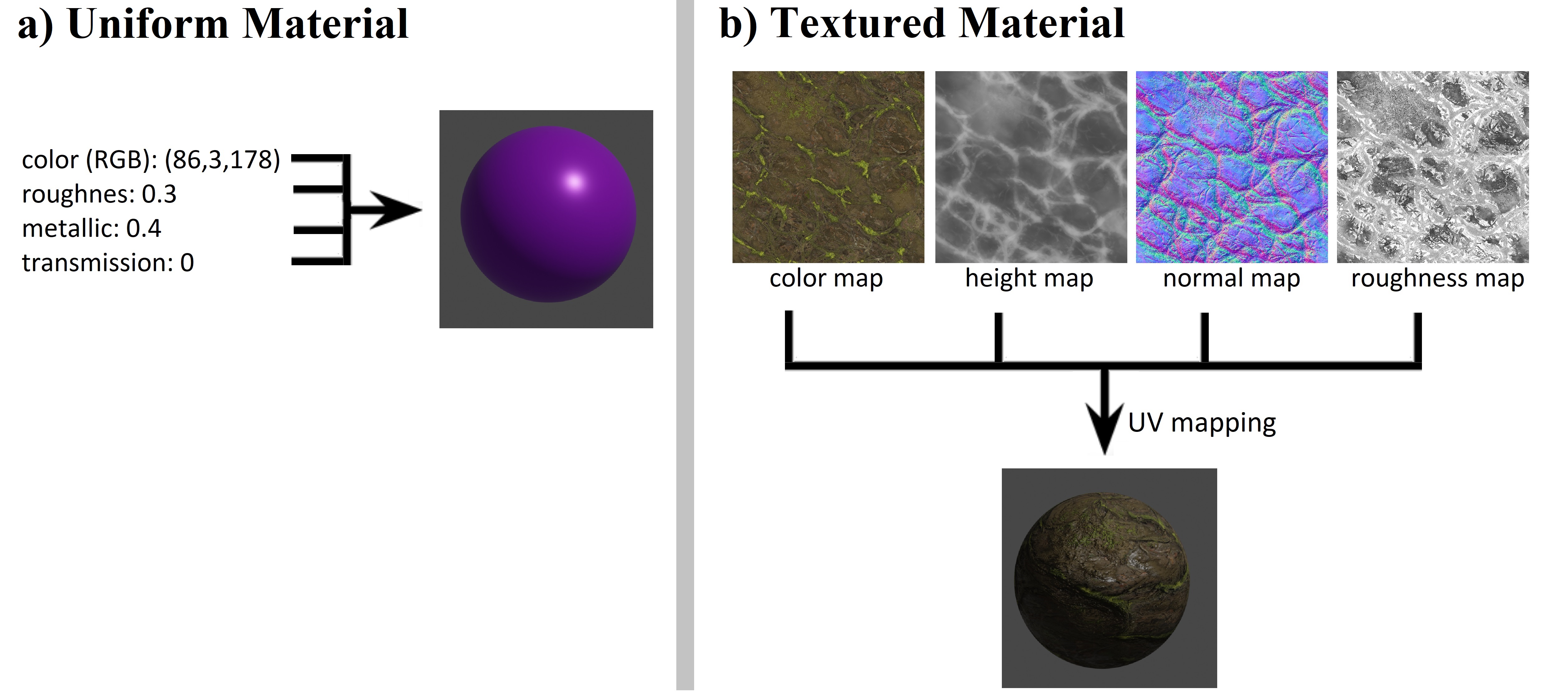}
   \caption{A material's visual appearance is controlled by several main properties (color, transparency, etc.). For uniform materials (BSDF), each property has a single value across the surface. Textured materials are represented by texture maps for each property. These maps are wrapped around the object (UV mapping) and provide properties for each point on the object's surface.}
   \label{fig:matCreation}
\end{figure}

\subsection{Materials Representation, Mixtures, and Gradual Transformations} \label{sec:mixtures}
The appearance of materials is mostly controlled by their surface-scattering properties. These properties are often referred to as bidirectional reflectance (BRDF) and the more general bidirectional scattering function\cite{bartell1981theory, asmail1991bidirectional} (BSDF, \Cref{fig:matCreation}a). Each surface point has a set of properties, such as roughness/transmittance/color, which determine the surface appearance. If the material is uniform, it could be represented as a set of values for each property across the entire surface. In Blender3D\cite{blender}, this is done using the BSDF node (\Cref{fig:matCreation}a).   Creating a random material could be done by setting a random value for each property. Mixing two such materials can be achieved by using a weighted average of the values of each property from each material: $P_{\text{mix}}=R \cdot P_a+(1-R) \cdot P_b$ where $P$ is the value of a property in materials $a$,$b$ and $R$ the mix ratio. Gradual transition between materials is achieved by creating different mixture ratios ($R=0,0.25,0.5,0.75,1$, \Cref{fig:dataSets}). Most materials in the world are not uniform and have unique textures, which means different properties for each point on the surface (\Cref{fig:matCreation}b). Such materials can be represented as spatially variable BRDF or SVBRDF (\cite{Pharr2016physically}). This means that instead of a single value for each property, we have a 2D texture map that represents the spatial distribution of this property on the surface (\Cref{fig:matCreation}b). This 2D map is then wrapped around the object using a process called UV mapping to give each surface point its property. Mixing two textured materials is achieved by a pixel-wise weighted average of two texture maps into a new texture map. In other words, each pixel in the texture map of a given property is the average of the corresponding pixels in the two materials that are mixed: $P_{\text{mix}}(u,v)=R \cdot P_a (u,v)+(1-R) \cdot  P_b (u,v)$. Where $P(u,v)$ is the property in surface position $u$,$v$. Gradual transition between materials $A$ and $B$ is again achieved by different mixtures ratios($R$). In order to increase variability, texture maps of different materials can be rotated and rescaled relative to the other material before mixing (but not relative to maps of the same material). Unlike uniform BSDF textures, it's not possible to create textures by assigning random values, as this will just create noise. We, therefore, got a large number of textured materials by downloading 4400 textures from free large-scale artists' repositories. These are highly diverse realistic materials scanned or generated for CGI purposes. To further increase this set, we mix two or more materials as described above, leading to millions of different textures. 
\begin{figure}[t]
  \centering
   \includegraphics[width=1.0\linewidth]{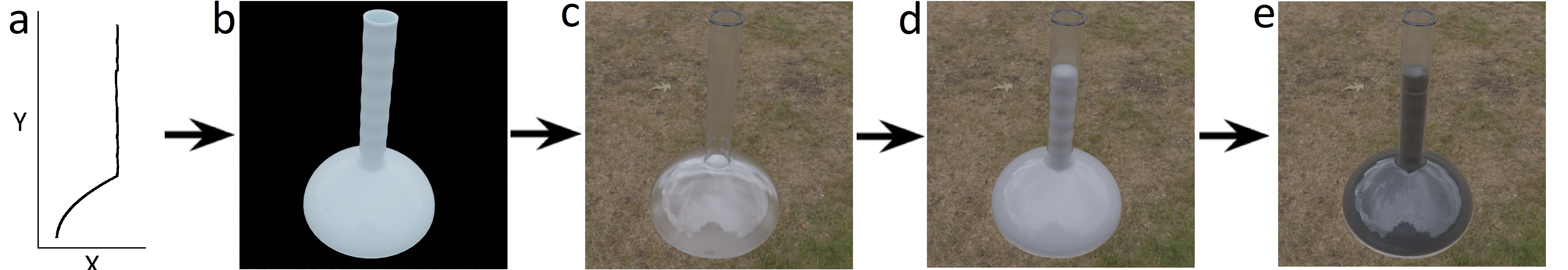}
   \caption{Procedurally generating material inside transparent containers. a) A random 2D curve is generated by combining random polynomial and trigonometric functions. b) The curve is used as a profile for the symmetric 3D object. c,d) The object is assigned random transparent materials and content objects. e) The content object is assigned a material.}
   \label{fig:transparent}
\end{figure}

\subsection{Materials in Transparent Vessels} \label{sec:transparentV}
\begin{figure}[t]
  \centering
   \includegraphics[width=0.9\linewidth]{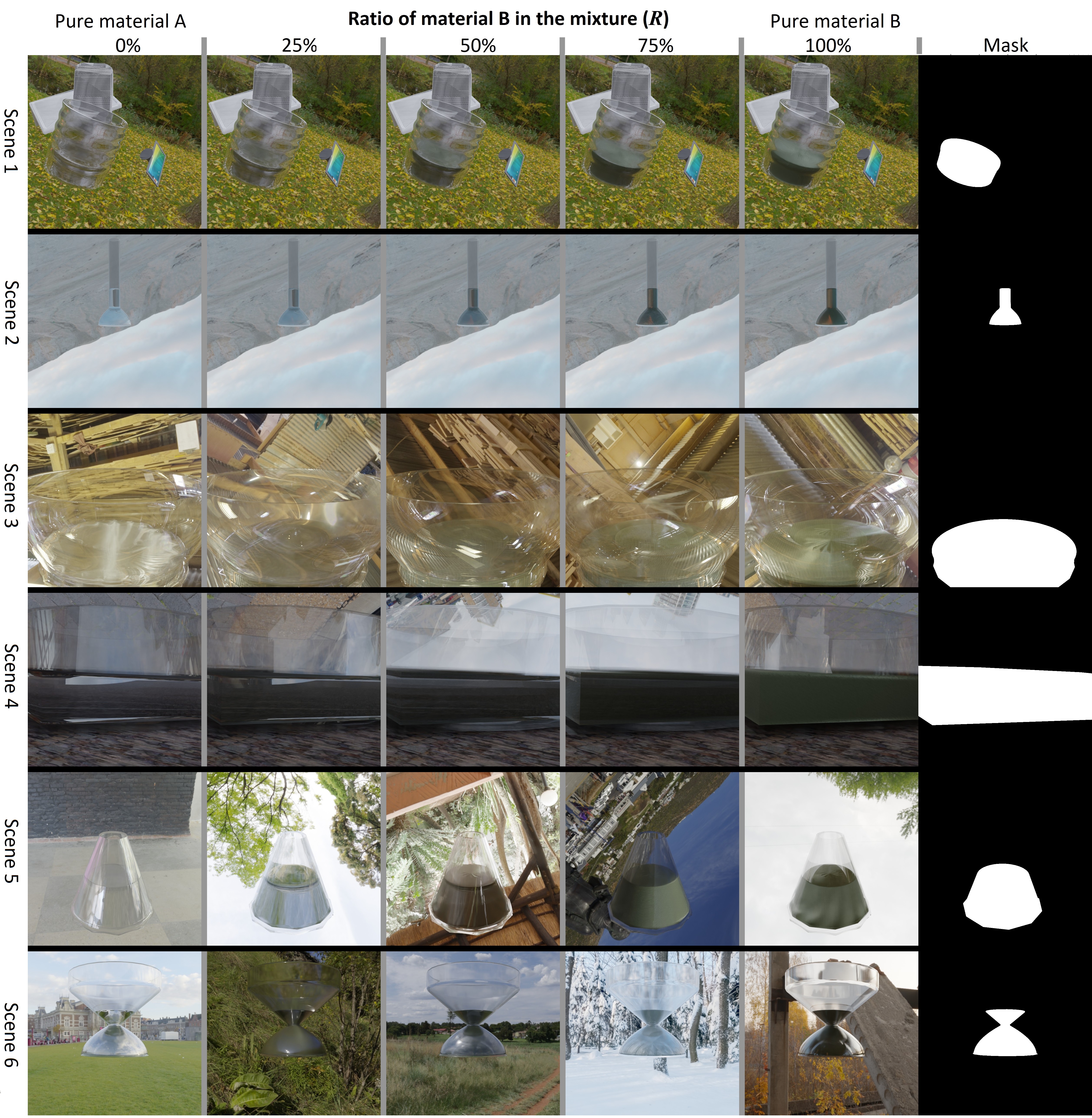}
   \caption{A Set of the MatSim dataset with materials inside transparent containers. The set structure is described in \Cref{sec:genDataset} and \Cref{fig:dataSets}. The vessel generation is described in \cref{sec:transparentV}}
   \label{fig:dataSets_ves}
\end{figure}

Liquids and other materials in kitchens, hospitals, and labs are usually handled inside transparent containers (glassware, flasks, tubes, etc.). To support these applications, we generated scenes in which we put the material inside a transparent container. The vessel object was procedurally generated. The curvature of the transparent vessel was generated by creating a random 2D curve (by randomly combining linear, polynomial, and trigonometric functions; \Cref{fig:transparent}). This curve was used as the profile of a symmetric vessel by creating a cylindrical (or other symmetrical) shape with the curve as the vertical profile (\Cref{fig:transparent}). In some cases, the shape was also randomly stretched to create more variability. The vessel object was assigned a random transparent material and a random thickness. The content of the vessel was either a random object loaded from ShapeNet or a mesh that filled the bottom part of the vessel (similar to static liquid or powder). As before, the content was assigned a random material. Otherwise, the dataset creation was the same as in \Cref{sec:genDataset}. Examples can be seen in \Cref{fig:dataSets_ves} (Appendix).

\section{Benchmark Creation}
Perhaps the most important part of dealing with the task of general material recognition is defining a proper benchmark that will cover the main aspects of this challenge. We define three main capabilities that we want to evaluate:
1) The ability of the network to visually identify unfamiliar materials in new environments using one or a few examples with no restrictions on the material or environment setting. 
2) The ability to identify transitions between material states (e.g., wet/dry, rotten/fresh) and fine-grained subcategories of materials (e.g., types of pasta).
3) The ability to identify materials regardless of the object they appear on (e.g., a cow made out of wood)\cite{barragan2022adversarially,hendrycks2021many}.
 We are not aware of a benchmark that covers any of these three tasks. Therefore, we created two benchmarks by manually taking pictures of materials in a variety of states.

\subsection{Set 1: Material transition states and subclasses}
The first test set is designed to test the hypothesis that the network is able to recognize the similarity between images of the same material, even when they are presented in transparent containers and different surroundings, while distinguishing between subcategories or different states of the same material. To do this, we collected a large number of images of materials from a wide range of fields (\Cref{fig:accExamples}); for example, we collected images of eggs in raw, cooked, and burned states and sidewalks in wet and dry states (\Cref{fig:accExamples}). For each state, we collected at least two images. We divided this test set into superclasses (e.g., eggs) and divided each superclass into subclasses (e.g., cooked/raw/burned; \Cref{fig:accExamples}). We tried to make different examples of the same material subclass appear in different environments as much as possible, making the recognition of the material based on the environment type less likely. In addition, for each image, we created a mask of the material region in the image. Note, however, that in this dataset, there will always be some correlation between materials and objects, i.e., liquids appear in glassware, and rotten textures appear on fruit. This set contained 416 images divided into 116 material types.

\subsection{Set 2: Uncorrelated materials and objects}
\begin{figure}[t]
  \centering
   \includegraphics[width=0.9\linewidth]{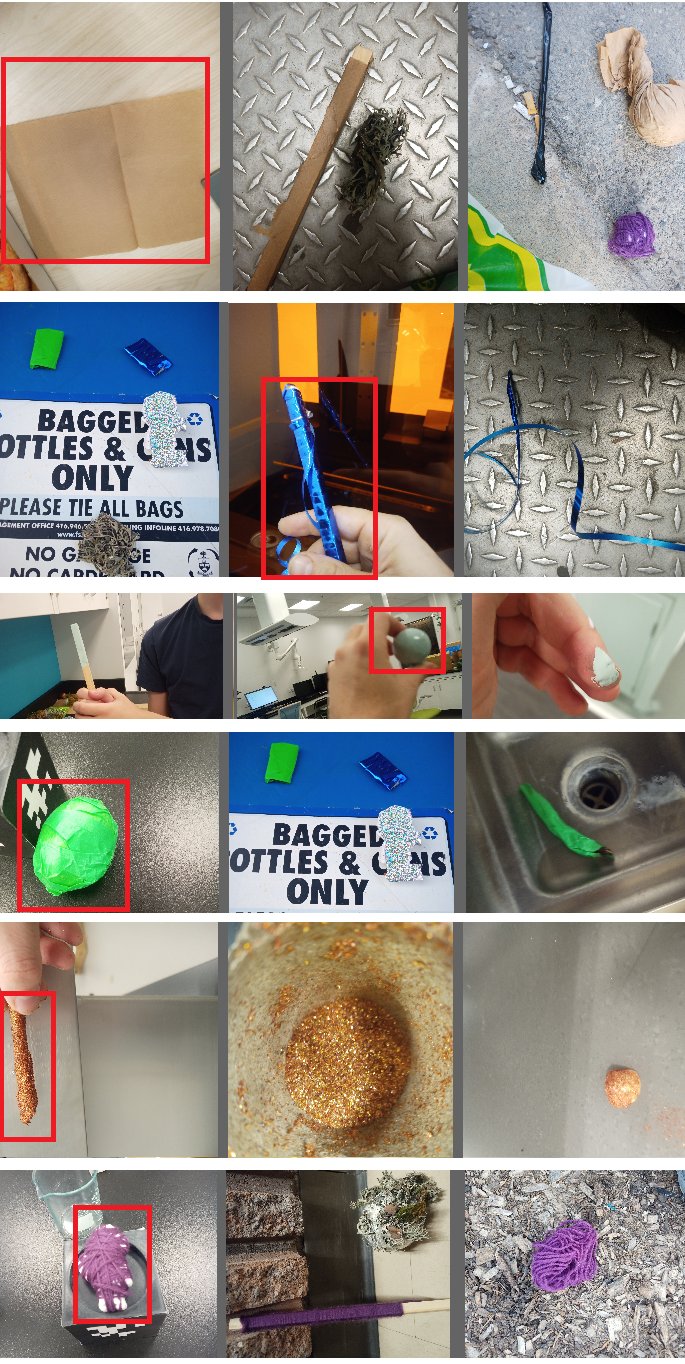}
   \caption{Samples from the second benchmark: uncorrelated materials and objects. Random materials cover random objects to avoid material and object correlation. Each row corresponds to one type of material. For clarity, a red square was used to mark the material in each row (This doesn't appear in the actual benchmark). Material masks are not shown but are available in the benchmark.}
   \label{fig:set2}
\end{figure}

The second test set is designed to test the hypothesis that the network is able to recognize the material regardless of the object on which it appears. Materials in the real world are strongly correlated to objects (trees made out of wood). Creating this set meant that we needed to generate our own objects. To achieve this, we collected a number of random objects and materials. The materials were ribbons and sheets that could be wrapped around the object or granular and fiber materials that could be scattered on the object's surface (\cref{fig:set2}, appendix). These materials were used to cover the surfaces of objects using glue for attachment. When creating this test set, we made sure that each material did not appear on the same object or in the same environment in more than one image, forcing the net to use only the material for recognition. This set contains 86 images in 16 material types.

 \subsection{Materials Classification Using a Descriptor}\label{sec:benchclass}
\begin{figure}[t]
  \centering
   \includegraphics[width=1.0\linewidth]{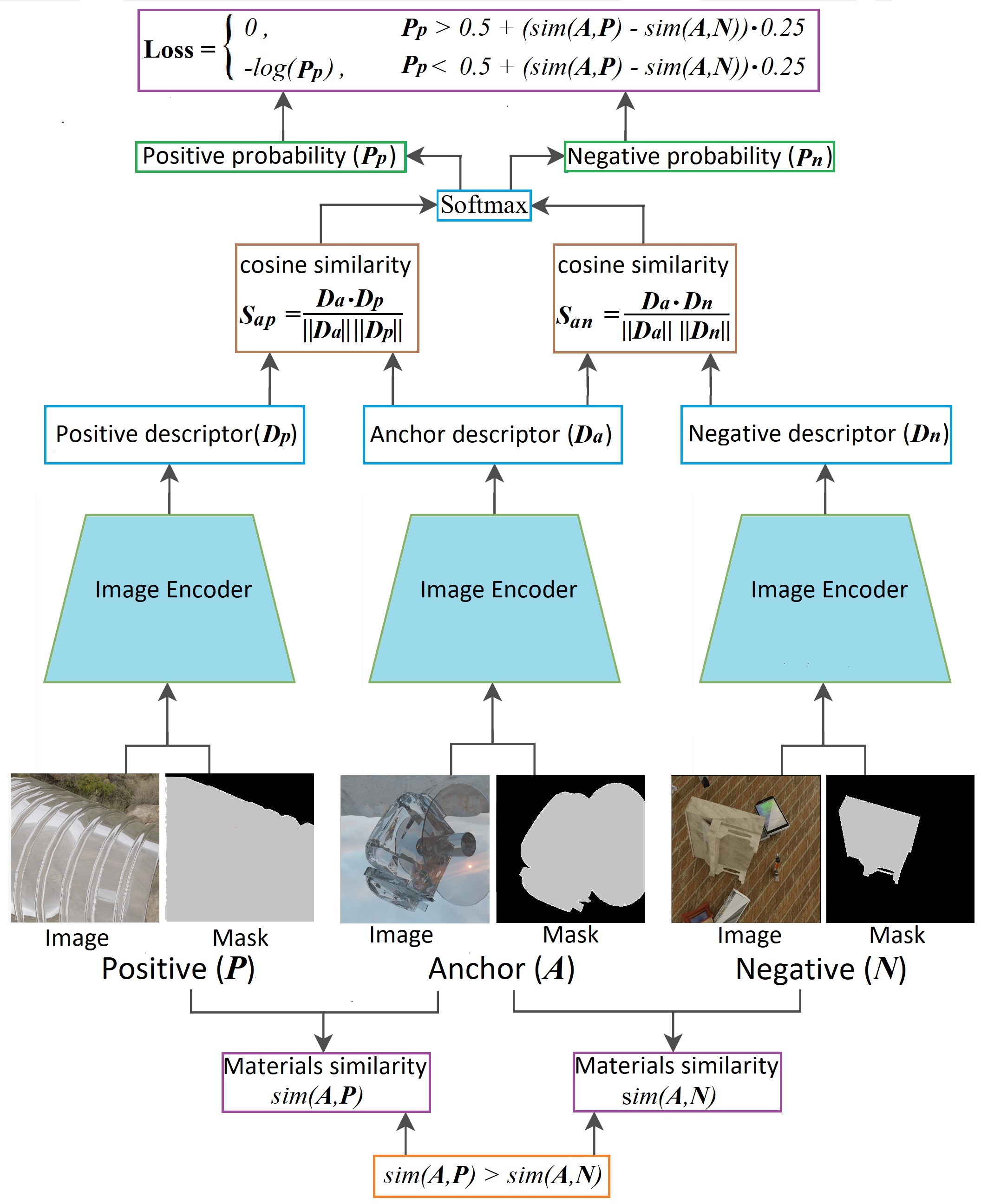}
   \caption{Training and loss function. The loss function is based on the cosine similarity with cross-entropy loss. For every three images in the batch, one image is defined as an anchor ($A$). A second image, specifically the image that has material more similar to the anchor, is defined as positive ($P$), and the third image is defined as negative ($N$). All images and the material masks are passed through the neural network to produce descriptors ($D_a$, $D_n$, $D_p$). The cosine similarity is calculated between the descriptor of the anchor and the positive and negative examples. These cosine similarities are the input for the softmax function, which returns the probability of a match between the anchor and the positive image ($P_{p}$). If this probability is below the threshold defined as $0.5+(sim(A, P)-sim(A, N))\cdot0.25$, we calculate the cross-entropy loss ($Loss = (-\log(P_p)$); otherwise, the loss is set to zero. $sim(A, P)=1-|R(A)-R(P)|$ is the similarity between the anchor ($A$) material and the positive material $P$,  $R(A)$ is the mixture ratio in material $A$ (\Cref{sec:mixtures}).}
   \label{fig:training}
\end{figure}

The MatSim dataset relies on visual similarity. It is not intended and is less suited for handling broad semantic material classes. General classes, like wood or plastic, may include a wide range of textures with little or no visual similarity between them. However, recent studies have shown that networks trained on self-similarity can often generalize effectively to broader semantic classes when provided with a large set of images \cite{van2020scan}. While not the primary focus of our work, we tested the MatSim-trained net and CLIP\cite{radford2021learning, schuhmann2022laion} on major material classification datasets. The evaluation process followed the same method used for other MatSim benchmarks (\Cref{sec:evmethod}), where each image in the dataset was matched to all other images, and if the best-matched image belonged to the same class, it was considered correctly classified. The accuracy for each class was calculated as the average for all images in that class, and the overall dataset accuracy was determined as the average accuracy across all classes. We examine the main datasets for materials classification: OpenSurface\cite{bell2015material, bell2013opensurfaces}, Flicker Materials (FMD)\cite{sharan2009material}, and Dense Material Segmentation(DMS)\cite{upchurch2022dense}. We use the material region/segment as the input mask (\Cref{fig:training}). For CLIP, we also tested the net ability to classify materials by matching the image to the text label(standard semantic classification).

\section{Evaluation Methods}\label{sec:evmethod}
For the evaluation of the network, we consider the standard Top-1 accuracy metric: Given an image of a material, we test the ability of the network to identify another image of the same material from a group of images. An example is correct if the highest similarity calculated by the network is  between the given image to another image of the same material subclass. We calculate two accuracy metrics. The first evaluation method measures the ability of the network to find another example of the same material from the set of all images of material the superclass (\Cref{fig:accExamples}). For example, the task could be to identify an image with spotted bananas from a set of images of bananas in various states (\Cref{fig:accExamples}, \Cref{tab:results} (subclass)). In the second test case, we still expect the network to find another example of the same material. However, we now present it with all the other images in the test set, including those of different superclasses. For example, the image of a spotted banana is compared to all images, including those of rust, rocks, cheese, etc. Note that in this case, it is much easier to use the object type to distinguish between materials (banana vs. not a banana), which gives the Clip network an advantage. For the second test set we used only the second approach. The results given in \Cref{tab:results}, \Cref{fig:accExamples}, are the average accuracy per class (All classes given equal weights).
\section{Training and Architecture} \label{sec:TrainingAndArch}

We examined several standard image encoder architectures and found that ConvNeXt\cite{liu2022convnet} gave the best results. Two modifications were made in ConvNeXt: First, the input of the network was changed from a standard three-channel (RGB) image to an image plus the mask of the region containing the material. This was done by simply adding the ROI mask as another layer to the RGB image and changing the first layer of the network so that it could receive a four-layer (R, G, B, mask) image instead of a three-layer (RGB) image. In addition, the final linear layer of ConvNeXt was changed to produce a 512-value vector (instead of 1000 ImageNet classes). This vector was then normalized using L2 normalization and used as the output descriptor. The training itself was done using the AdamW optimizer for ten epochs (200000 steps with a batch size of 12) using a single RTX3090. This took around three days per training session. We used the ConvNeXt-base model pretrained on ImageNet.

\subsection{General Loss Function}
The loss function is described in \Cref{fig:training}.  Similar to recent works, we use the cosine similarity combined with the cross-entropy loss for training \cite{chen2020simple}. An important aspect of our dataset is that the levels of similarity between materials are continuous and can have any value between zero and one (depending on the mixture; \Cref{fig:dataSets}). This was accounted for by using a semi-hard loss with a threshold depending on the similarity level. We found that a semi-hard loss, in which only loss terms that are below some threshold are used, works better than a hard loss (in which all loss terms are used). The threshold, in this case, was controlled by the difference in the similarities between the anchor material and the positive and negative examples.

\subsection{Material Similarity} \label{sec:materialSim}
We define the real similarity of the two materials as the difference between their mixture ratios ($R$, \Cref{fig:dataSets}). As described in \Cref{sec:mixtures}, each material in a set is a linear mixture between two materials $A$ and $B$. We define this ratio as $R(0 < R < 1)$. The similarity between the materials in the two images ($i_1$ and $i_2$) is defined as: \begin{equation*}sim(i_1,i_2)=1-|R(i_1)-R(i_2)|\end{equation*} Where $R(i_1)$ is the mixture ratio in image $i_1$ (\Cref{sec:mixtures}). Assuming both images are from the same set.

\subsection{Predicted Material Similarity} \label{sec:materialPrediction}
The predicted material similarity between two images, $A$ and $N$, is calculated by first passing the images and their corresponding masks through the neural network and predicting the descriptors $D_a$ and $D_n$, which are then normalized using L2 normalization. The predicted similarity is the cosine similarity between the two descriptors ($S_{an}$, \Cref{fig:training}). 

\subsection{Loss Calculation}
The training was performed as shown in \Cref{fig:training}: For each batch, we randomly selected a single set (\Cref{sec:genDataset}) and sampled 12 random images from this set. The loss for every three images in the batch was calculated separately using the procedure shown in \Cref{fig:training}. One of the three images was chosen as an anchor ($A$). The similarity between the material in the anchor image and the other two images was calculated as described in \Cref{sec:materialSim}. The image that was more similar to the anchor was defined as positive ($P$), and the other image was defined as negative ($N$); hence, $sim(A, P) > sim(A, N)$.
If the similarities of the anchor to the two images were equal $(sim(A, P) = sim(A, N))$, the loss for this triplet was set to zero. Next, each of the three images and their material masks were passed through the neural net to predict the material descriptors (\Cref{sec:TrainingAndArch}, \Cref{fig:training}). The cosine similarities between the descriptors of the anchor and the negative and positive images were calculated as described in \Cref{sec:materialPrediction}. These similarities were then passed to a softmax function to calculate the probability of a match between the anchor and the positive image:  \begin{equation}
    P_p = \frac{e^{S_{ap}/t}} {e^{S_{ap}/t} + e^{S_{an}/t}}.
\end{equation}
$S_{ap}$ and $S_{an}$ are the cosine similarity between the anchor ($A$) and the negative($N$) and positive($P$) descriptors, respectively, $P_p$ is the probability for a match between the anchor and the positive image, and $t$ is a constant (0.2 in our case). We then calculate the semi-hard loss using the following condition:
If $P_p > 0.5+(sim(A, P)-sim(A, N))\cdot0.25$, the loss is set to zero; otherwise, we calculate the standard cross-entropy loss $(Loss = -log(P_p))$. 


\section{Testing Clip Models}
The Clip neural network has achieved state-of-the-art (SOTA) results on a large number of image recognition benchmarks and is considered the top general one-shot net to date \cite{schuhmann2022laion,radford2021learning}. Similar to our model, Clip predicts image descriptors as a vector, which can then be used to find similarities between the image and other images or texts. Unlike our model, Clip H14 was trained by comparing 2 billion real photos to their corresponding text captions. There are two limitations to using Clip for this task. The first is that the Clip descriptor is not focused on materials and might contain information regarding the object and environment. Second, Clip receives just an image without an attention mask. Hence, it is harder to point to a specific region of the image where the target material is. To solve this issue, we tried a few methods: The first involves cropping the region of the image around the target material and using it as the input. The second involves masking the image region outside the material (by setting it to black), and the third involves combining masking and cropping. All of these approaches gave a significant improvement compared to using the image as is. Cropping without masking gave the best results. We tested all the pretrained Clip versions available and found that ViT-B32 gave the best results for open A.I models (\Cref{tab:results}), but the recently published Open Clip H/14 significantly outperforms all other Clip Models.

\section{Results}

\begin{table}
  \centering
  \begin{tabular}{@{}lccc@{}}
    \toprule
    Method & Set1 Subclass & Set1 All & Set2 \\
    \midrule
    Random        & 0.30 & 0.006 & 0.07 \\
    \midrule
    MatSim     & 0.71 & 0.56 & 0.73 \\
    MatSim+C   & 0.77 & 0.56 & 0.85 \\
    MatSim+M   & \textbf{0.78} & 0.56 & \textbf{0.91} \\
    MatSim+C+M & 0.72 & \textbf{0.61} & 0.85 \\
    \midrule
    Open CLIP H14     & 0.55 & 0.44 & 0.47 \\
    Open CLIP H14+C   & 0.67 & 0.52 & 0.77 \\
    Open CLIP H14+M   & 0.59 & 0.40 & 0.53 \\
    Open CLIP H14+C+M & 0.66 & 0.52 & 0.67 \\
    \midrule
    CLIP B32     & 0.51 & 0.32 & 0.44 \\
    CLIP B32+C   & 0.56 & 0.38 & 0.49 \\
    CLIP B32+M   & 0.56 & 0.28 & 0.37 \\
    CLIP B32+C+M & 0.56 & 0.35 & 0.56 \\
    \midrule
    Human Similarity    & 0.41 & 0.13 & 0.22 \\
    Human Similarity+C   & 0.60 & 0.20 & 0.41 \\
    Human Similarity+M   & 0.55 & 0.15 & 0.27 \\
    Human Similarity+C+M & 0.55 & 0.16 & 0.42\\ 
    \midrule
    MatSim+M No Augmentation    & 0.66 & 0.45 & 0.61 \\
    MatSim+M No vessels   & 0.72 & 0.52 & 0.61 \\
    MatSim+M No mixtures  & 0.74 & 0.53 & 0.72 \\
    \bottomrule
  \end{tabular}
  \caption{Results. $+C$ indicates cropping, $+M$ indicates masking. Random stands for random matching. Human Similarity refers to the net trained on human-annotated material similarity metrics\cite{lagunas2019similarity}. No vessels refer to a net trained only on the objects part of the MatSim dataset. No mixtures refer to a net trained on the MatSim with no materials mixtures.}
  \label{tab:results}
\end{table}

\begin{table}
  \centering
  \begin{tabular}{@{}lccc@{}}
    \toprule
    Method & OpenSurface & DMS & FMD \\
    \midrule
    Random             & 0.03 & 0.019 & 0.09 \\
    MatSim+M           & \textbf{0.33} & \textbf{0.29}  & 0.66 \\
    Open CLIP H14+C    & 0.32 & 0.14  & \textbf{0.82} \\
    Human Similarity+C & 0.07 & 0.04  & 0.20 \\
    \midrule
    Open CLIP H14+C Semantic    & 0.30 & 0.19 & 0.72\\
    Semantic Random    & 0.03 & 0.018 & 0.1\\

    \bottomrule
  \end{tabular}
  \caption{Results for Material Classification on Standard Benchmarks(\Cref{sec:benchclass}). Random stands for random matching. Semantic stand for matching the image to the text labels. $+C$ indicates cropping, $+M$ indicates masking. The attention mode for each net (masking/cropping) is the one that gives the best results.}
  \label{tab:resultsclass}
\end{table}

As seen in \Cref{fig:accExamples} and \Cref{tab:results}, the MatSim trained net performed well across almost all types of materials and environments. The MatSim net significantly outperforms the best Clip model on all test sets (\Cref{tab:results}). This suggests that despite the fact that the MatSim network was trained only on simulated data and visual self-similarity, it learned to generalize to unfamiliar real-world material states. Both Clip and MatSim significantly outperformed net trained on human-annotated material similarity metrics\cite{lagunas2019similarity}. All approaches significantly outperformed random matching (\Cref{tab:results}). Both MatSim net and Clip performed well on Set 2 despite this set having no correlation between materials and objects. This supports the hypothesis that the recognition is done only based on material features and not correlated properties.

 \subsection{Results for General Material Classification}
 The MatSim visual similarity approach is not intended for the more general semantic classes, which are associated more by names and less by appearance. However, as can be seen from \Cref{tab:resultsclass}, both CLIP and MatSim net gave relatively good results for matching images of the same general classes (plastic, glass, wood…). For the  DMS\cite{upchurch2022dense} and OpenSurface\cite{bell2015material, bell2013opensurfaces}  datasets, the net trained on MatSim outperformed CLIP (\Cref{tab:resultsclass}). This could be attributed to the larger number of images and classes, making it more likely for similar textures to occur in each class, while the more diverse class set also makes semantic-guided classification harder.   CLIP performed better than MatSim net on the FMD benchmark\cite{sharan2009material}. This can be explained by the fact that this dataset contains a small set of general classes and a relatively small number of images (100) per class, where each class can contain a large number of different textures, making semantic knowledge of the material class a major advantage, while limiting the effectivity of relying only on textures visual similarity.  The net trained using human-assigned materials similarity data\cite{lagunas2019similarity} performed by far the worst on all datasets but still way above random.

\subsection{Semantic Material Classification Using CLIP}
We also tested the ability of the CLIP text/image embedding model\cite{radford2021learning, schuhmann2022laion} to classify materials by their semantic text labels. This was done by matching materials images to their text labels (classes), as well as all other Labels in the dataset. A material was correctly classified if its best match was for the correct class label. As seen from  \Cref{tab:resultsclass}, Clip H14 performs far better than random on this task and near image-to-image-based matching accuracy.
Clip achieved 30\% accuracy on the semantic classification of the OpenSurface dataset\cite{bell2015material, bell2013opensurfaces}. This is well below the 50\% accuracy achieved by a small Resnet50 trained specifically on the same dataset\cite{eppel2018classifying}. This contrasts CLIP results on major datasets like ImageNet, where Clip performed at the same level as a much larger Resnet101 trained specifically on this dataset\cite{radford2021learning, schuhmann2022laion}. This implies that the CLIP model and training approach might be less effective for material understanding compared to other tasks. Alternatively, it might be that Clip cannot focus on the specific region of the material in the image and that the cropping/masking methods tested for this purpose are ineffective (Unlike ImageNet, where each image is dedicated to one class, materials datasets usually contain a number of materials and their regions).

\subsection{Dataset Components and Their Effects}
The results in \Cref{tab:results} demonstrate that Augmentation, Training for materials inside transparent vessels, and training with a gradual transition between materials are all vital for the performance of the MatSim network. Eliminating any of these resulted in a considerable decline in performance, but the network still performed better than CLIP even without these aspects. Augmentation seems to be the most crucial factor. However, this is likely applicable to most image-based neural networks.

\subsection{Comparing Masking and Cropping}
Focusing the network on the material region of the image by cropping the material region and using it as the input image seems to improve the accuracy of both Clip and the MatSim network (\Cref{tab:results}). Masking the background by blacking out the background pixels seems to improve the results for the MatSim network but proves to be less effective than cropping for Clip. The fact MatSim net is affected by cropping and masking is surprising  because the MatSim network already receives the region of the material as input (\Cref{sec:TrainingAndArch}), so ideally, it should be able to focus on this region without either cropping or masking. Not surprisingly, Clip with no cropping or masking has the worst performance since it has no inherent way to focus on the material region.

\section{Conclusion}

This work introduces a new approach, dataset, and benchmark for general one-shot material recognition, not limited to specific material types or settings. It tackles several challenges related to material recognition, including material states, subclasses, mixtures, gradual state transition, and recognition of materials inside transparent containers. One-shot recognition of materials has been mostly overlooked in image recognition research, but it is critical for understanding different aspects of the world and has numerous applications, ranging from material science and chemistry to cooking and agriculture. Our findings demonstrate that a network trained on self-similarity using diverse synthetic data can recognize almost every material state using just one or a few examples, regardless of the environment, outperforming large-scale models trained on human-generated semantic labeling (CLIP) and similarity metrics based on human perception. We hope that the dataset and benchmark will pave the way for testing and developing new techniques in this emerging field.

{\small
\bibliographystyle{ieee_fullname}
\bibliography{egbib}
}
\section{Appendix: Building the MatSim Dataset} \label{sec:Appendix}

\subsection{Setting objects in the scene} \label{sec:apsetobjinscene}
Objects for the dataset were taken from the ShapeNet core2 dataset\cite{chang2015shapenet}, which has 56000 3D model objects with over 200 categories. Each object is loaded, randomly scaled, and rotated, and its original materials are removed. In addition, overlapping faces on the surface are removed. 
\subsection{UV mapping} \label{sec:apuvmap}
Since the SVBRDF (PBR) materials are given as 2D texture maps, assigning a material to a 3D object involves wrapping these 2D maps around the object's surface. Wrapping 2D maps around complex 3D objects (UV mapping) is a rather complex problem. Blender 3.1 contains some automatic wrapping techniques, which we use. A major problem that we encounter is that many objects contain overlapping faces (surfaces), which causes the textures to be wrapped around the same areas several times, causing strong unrealistic visual artifacts. We managed to solve this by removing vertexes and faces that were too close to each other using the remove overlapping vertexes tool of Blender. This led to the slight deformation of some objects' shapes. The wrapping itself depends on various parameters, such as the relative scale, orientation, and translation of the texture map relative to the object, as well as the coordinate system. We randomize all of these parameters between each image to achieve maximum variability. 
\subsection{Background Illumination and HDRI} \label{sec:aphdri}
The environment and illumination greatly affect a material's appearance. The main method used by the CGI community to control the background and illumination involves high dynamic range images (HDRIs). These are panoramic images that wrap around the scene and provide a 360 illumination. The range of intensities of light is from 0–6550, compared to 0–255 in standard images. This allows HDRIs to represent a much broader range of illumination. We downloaded over 500 HDRIs from the polyhaven repository\cite{polyhaven}. These HDRIs are highly diverse scenes covering both daytime and nighttime conditions in a large number of indoor and outdoor environments. These HDRIs were used to add both illumination and background to the scenes. To increase diversity, the HDRIs were randomly rotated and scaled, and their intensities were randomly increased or decreased for each scene.
\subsection{Adding Ground Plane and Background Objects} \label{sec:apglane}
Random objects were scattered in the scene to make the environment more diverse and add shadows and back reflections. This was done by loading random objects from the ShapeNet dataset and randomly scaling, rotating, and positioning them in the scenes. In addition, a ground plane was generated by adding a flat plane below the object and assigning a random PBR material to it.
\subsection{Limitation of the Synthethic Dataset}
A major limitation of the dataset is that phase transitions between materials are often not the average between the two states; instead, they are either a completely different state (the transition between wood and coal is fire), or they are not uniformly distributed, like the nucleation of ice in water. Both cases are not completely addressed by the dataset and require further work.

\subsection{Data Augmentation}
A major problem with using only simulated data for learning is the fact that modern cameras significantly modify an image's appearance by smoothing and adjusting colors. To account for this, we use extensive augmentation for the image during training. This includes random Gaussian smoothing, the darkening or brightening of the image, partial or complete decoloring (to greyscale), and noise addition. Each of these augmentations was applied for about 10\%-20\% of the image regardless of whether other augmentations were used. Resizing, cropping, and flipping was done for almost every image. Combined, this led to an improvement in accuracy of over 10\%-30\% (compared to the non-augmented version).
\subsection{Data sampling}
Sampling is done by choosing one set for each batch and, from this set, choosing 12 random images (with no repetition). Sets with materials inside and outside transparent vessels were selected with equal probability.

\subsection{CGI Assets From Artist Repositories}
One of the main challenges in creating the MatSim dataset was collecting large and diverse enough examples of backgrounds, objects, and materials. Large-scale repositories of assets that serve the CGI artist community have been available for a while but remain almost unutilized by the machine-learning community. Utilizing these repositories made it possible for us to create a wide range of scenes on a large scale and with sufficient diversity; this would have been very difficult if we had created these data ourselves. The main three repositories used for this work were AmbientCG, CGBookCase and TextureBox for materials,  HDRI Haven for environments and light, and ShapeNet for objects.

\subsection{Blender Render Setting and Hardware.}
The dataset was rendered using Blender 3.1, with  CYCLEs rendering, 120 rendering cycles per frame per, images were created with 800X800 resolution and noise reduction mode (smoothing).  Nvidia RTX 3090

\subsection{Realism vs. Generality in the Dataset Creation}
There are two ways to go about generating a synthetic dataset. One approach is to make the dataset as realistic as possible by maximizing the similarity of the generated data to the real world. The other approach is to maximize variability and make the scenes as diverse as possible, even at the cost of realism. In this work, we prefer variability over realism, which means we assign random materials to random objects and set them in random environments. For example, an image in the dataset may contain a car made out of wood on the ground made of metal in a forest surrounded by random objects. This makes the dataset easier to generate and helps the network achieve a much higher level of generalization and identify materials regardless of the environment or the object they appear on.

\end{document}